\pdfoutput=1

\documentclass[11pt]{article}

\usepackage{ACL2023}

\usepackage{times}
\usepackage{latexsym}

\usepackage[T1]{fontenc}

\usepackage[utf8]{inputenc}

\usepackage{microtype}

\usepackage{inconsolata}

\usepackage{graphicx}
\usepackage{algorithm}
\usepackage{algorithmic}
\usepackage{arydshln}
\usepackage{comment}
\usepackage{multirow}
\usepackage{booktabs}
\usepackage{subfigure}

\title{Do You Hear The People Sing? Key Point Analysis via Iterative Clustering and Abstractive Summarisation}

 \author{Hao Li$^\clubsuit$ Viktor Schlegel$^{\diamondsuit \clubsuit}$ Riza Batista-Navarro$^{\clubsuit 
 }$ \and Goran Nenadic$^\clubsuit$ \\
         $^\clubsuit$Department of Computer Science, University of Manchester, United Kingdom \\
         $^\diamondsuit$ASUS Intelligent Cloud Services (AICS), Singapore\\
         \texttt{\{hao.li-2, riza.batista, gnenadic\}@manchester.ac.uk}\\
         \texttt{viktor\_schlegel@asus.com} \\
         }

\begin{document}

\maketitle

\begin{abstract}
Argument summarisation is a promising but currently under-explored field. 
Recent work has aimed to provide textual summaries in the form of concise and salient short texts, i.e., key points (KPs), in a task known as  Key Point Analysis (KPA). 
One of the main challenges in KPA is finding high-quality key point candidates from dozens of arguments even in a small corpus.  
Furthermore, evaluating key points is crucial in ensuring that the automatically generated summaries are useful. 
Although automatic methods for evaluating summarisation have considerably advanced over the years, they mainly focus on sentence-level comparison, making it difficult to measure the quality of a summary (a set of KPs) as a whole. Aggravating this problem is the fact that human evaluation is costly and unreproducible. 
To address the above issues, we propose a two-step abstractive summarisation framework based on neural topic modelling with an iterative clustering procedure, to generate key points which are aligned with how humans identify key points. Our experiments show that our framework advances the state of the art in KPA, with performance improvement of up to 14 (absolute) percentage points, in terms of both ROUGE and our own proposed evaluation metrics\footnote{Our code can be found on Github: \url{https://github.com/HarrywillDr/keypoint-Analysis}}. Furthermore, we evaluate the generated summaries using a novel set-based evaluation toolkit. Our quantitative analysis demonstrates the effectiveness of our proposed evaluation metrics in assessing the quality of generated KPs. Human evaluation further demonstrates the advantages of our approach and validates that our proposed evaluation metric is more consistent with human judgment than ROUGE scores.
\end{abstract}

\section{Introduction}

Automated summarisation of salient arguments from texts is a long-standing problem, which has attracted a lot of research interest in the last decade. Early efforts proposed to tackle argument summarisation as a clustering task, implicitly expressing the main idea based on different notions of relatedness, such as argument facets \citep{DBLP:conf/sigdial/MisraEW16}, similarity \citep{DBLP:conf/acl/ReimersSBDSG19} and frames \citep{DBLP:conf/emnlp/AjjourAWS19}. However, they do not create easy-to-understand summaries from clusters, which leads to unmitigated challenges in comprehensively navigating the overwhelming wealth of information available in online textual content.

Recent trends aim to alleviate this problem by summarising a large collection of arguments in the form of a set of concise sentences that describe the collection at a high-level---these sentences are called \emph{key points} (KPs). This approach was first proposed by \citet{DBLP:conf/acl/Bar-HaimEFKLS20}, consisting of two subtasks, namely, \emph{key point generation} (selecting key point arguments from the corpus) and \emph{key point matching} (matching arguments to these key points). Later work applied it across different  domains \citep{DBLP:conf/emnlp/Bar-HaimKEFLS20}, for example for product/business reviews \citep{DBLP:conf/acl/Bar-HaimEKFS20}. While this seminal work advanced the state of the art in argument summarisation, a bottleneck is the lack of large-scale datasets. A common limitation of such an extractive summarisation method, is that it is difficult to select candidates that concisely capture the main idea in the corpus from dozens of arguments. Although \citet{DBLP:conf/acl/Bar-HaimEKFS20} suggested extracting key point candidates from the broader domain (e.g. selecting key point candidates from restaurant or hotel reviews when the topic is ``\emph{whether the food served is tasty}'') to overcome this fundamental limitation, it is impractical to assume that such data will always be available for selection. An alternative, under-explored line of work casts the problem of finding suitable key points as \emph{abstractive summarisation}. Research work in this direction aims to generate key points for each given argument, without summarising multiple of them \citep{DBLP:conf/argmining/KapadnisPPMN21}. As such, their approach rephrases existing arguments rather than summarising them.

One possible reason for key point generation being under-explored, is the lack of reliable automated evaluation methods for generated summaries. Established evaluation metrics such as ROUGE \citep{lin2004rouge} and BLEU \citep{papineni2002bleu} rely on the $n$-gram overlap between candidate and reference sentences, but are not concerned with the \emph{semantic similarity} of predictions and gold-standard (reference) data. Recent trends consider automated evaluation as different tasks, including unsupervised matching \citep{DBLP:conf/emnlp/ZhaoPLGME19,DBLP:conf/iclr/ZhangKWWA20}, supervised regression \citep{DBLP:conf/acl/SellamDP20}, ranking \citep{DBLP:conf/emnlp/ReiSFL20}, and text generation \citep{DBLP:conf/nips/YuanNL21}. While these approaches model the semantic similarity between prediction and reference, they are limited to per-sentence evaluation. However, this is likely insufficient to evaluate the quality of multiple generated key point summaries as a whole. For instance, the two key points \emph{``Government regulation of social media contradicts basic rights''} and \emph{``It would be a coercion to freedom of opinion''} essentially contain the same information as the reference \emph{``Social media regulation harms freedom of speech and other democratic rights''}, but individually contain different pieces of information. 

\begin{figure}[t!]
    \centering
    \includegraphics[width=\linewidth, scale = 1]{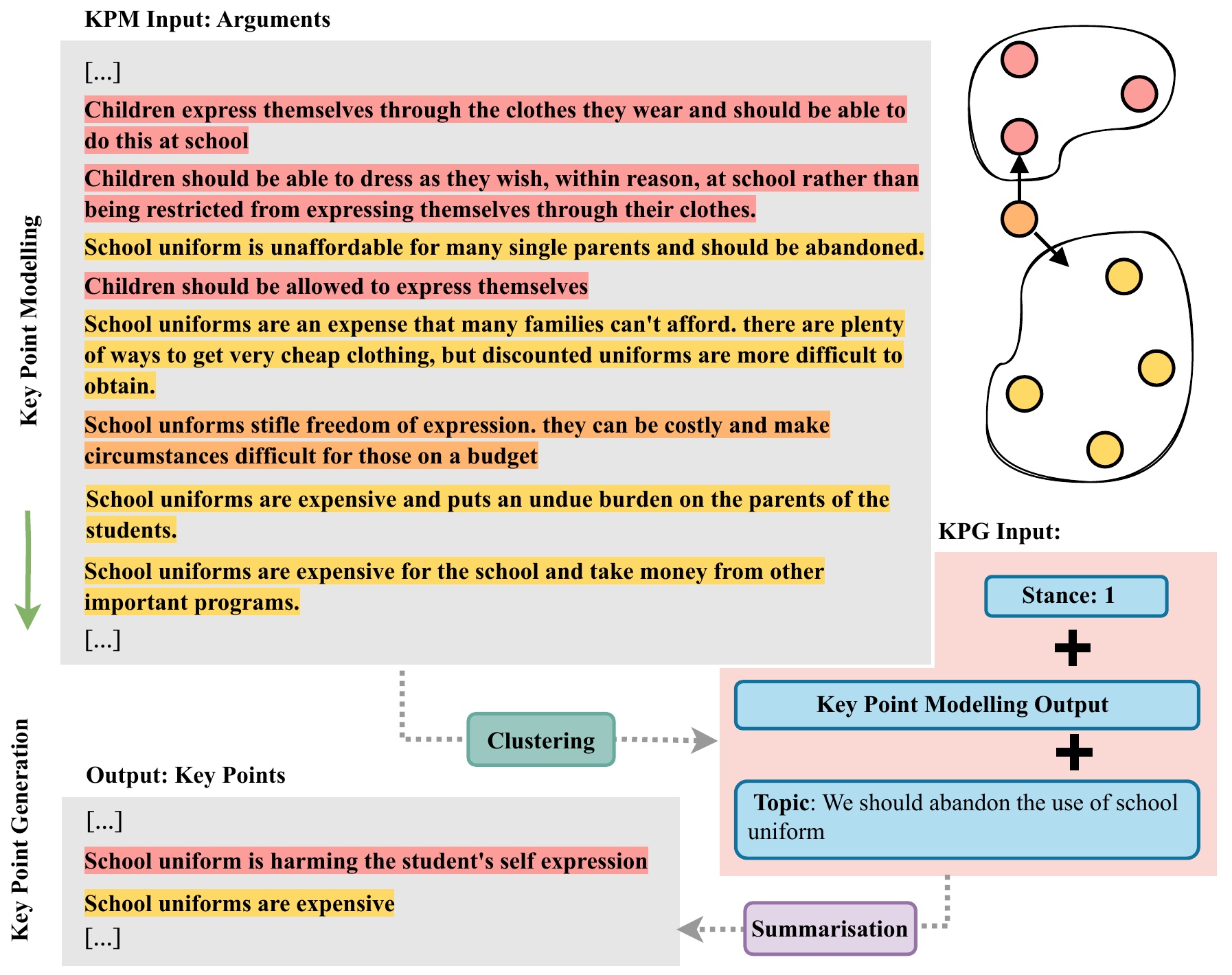}
    \caption{Visual depiction of our proposed framework. Colours illustrate the correspondences between arguments and key points. Nodes in \colorbox{orange}{orange} represent many-to-many matches, i.e., key points that are shared between both clusters. The input for key point generation (KPG) is composed of a single cluster from Key point Modelling (KPM) with its corresponding stance and topic. Key point importance is measured by the size of the clusters. For example, \textsc{Key Point: School uniforms are expensive} (\colorbox{yellow}{yellow}) has an importance of 5 (including the \colorbox{orange}{argument} that belongs to both clusters).} 
    \label{overview}
\end{figure}

In this work, we propose a novel framework for generative key point analysis, in order to reduce the reliance on large, high-quality annotated datasets. Compared to  currently established frameworks \citep{DBLP:conf/acl/Bar-HaimEFKLS20,DBLP:conf/emnlp/Bar-HaimKEFLS20}, we propose a novel two-step abstractive summarisation framework. Our approach first clusters semantically similar arguments using a neural topic modelling approach with an iterative clustering procedure. It then leverages a pre-trained language model to generate a set of concise key points. Our approach establishes new state-of-the-art results on an existing KPA benchmark without additional annotated data. Results of our evaluation suggest that ROUGE scores that assess generated key points against gold standard ones do not necessarily correlate with how well the key points represent the whole corpus. The novel \emph{set-based} evaluation metric that we propose, aims to address this.

Overall, the main contributions of this work are as follows:
    We propose a novel framework for key point analysis, depicted in Figure~\ref{overview}, which significantly outperforms the state of the art, even when optimised on a limited number of manually annotated arguments and key points. The framework improves upon an existing neural topic modelling approach with a semantic similarity-based procedure. Compared to previous work, it allows for better handling of outliers, which helps to extract topic representations accurately. Furthermore, we propose a toolkit for automated summary evaluation taking into account semantic similarity. While previous approaches concentrated on sentence-level comparisons, we focus on corpus-level evaluation. 
    
        
    

\section{Related work}
\textbf{Argument Summarisation}: 
The field of argument summarisation has developed considerably in recent years. \citet{DBLP:conf/coling/SyedBKKSP20,DBLP:conf/acl/SyedKAWP21} used an attention-based neural network to construct concise and fluent summaries of opinions in news editorials or social media. Alshomary et al. \citeyearpar{DBLP:conf/sigir/AlshomaryDW20}, focussing on web search, introduced an unsupervised extractive summarisation approach to generate argument snippets representing the key claim and reason. All of these efforts tackled \emph{single} document summarisation where only one argumentative text is summarised at a time. The earliest multi-document summarisation work attempted to summarise argumentative discussions in online debates by extracting summaries in the form of salient ``points'', where a point is a verb and its syntactic arguments \citep{DBLP:conf/acl/EganSW16}. However, their approach relies on lexical features that make it difficult to capture variability in claims that share the same meaning but are expressed differently.  
The work of \citet{DBLP:conf/emnlp/AjjourAWS19} and \citet{DBLP:conf/acl/ReimersSBDSG19} aimed to cluster semantically similar arguments. 
However, these efforts did not attempt to summarise these clusters, hence main points in the corpus remained implicit. 
Recent work proposed Key Point Analysis, which aims to extract salient points from a corpus of arguments, providing a textual and quantitative view of the data \citep{DBLP:conf/acl/Bar-HaimEFKLS20, DBLP:conf/emnlp/Bar-HaimKEFLS20}. \citet{DBLP:conf/argmining/AlshomaryGSHSCP21} contributed to the development of this framework by proposing a graph-based extractive summarisation approach. One common limitation of extractive summarisation methods, however, is that it is difficult to select key point candidates that truly capture salient points from dozens of arguments . \citet{DBLP:conf/argmining/KapadnisPPMN21} used an abstractive summarisation method, where each single argument and its topic were used as input in order to generate summaries.
A set of sentences which have the highest scores based on ROUGE \citep{lin2004rouge} ranking, is then selected as key points. 
However, in practice this is not feasible as the computation of ROUGE scores requires the availability of gold standard key points.

\textbf{Automatic Evaluation of Generated Summaries}: Most of the current work relies on human-centric evaluation methods \citep{DBLP:conf/argmining/AlshomaryGSHSCP21,DBLP:conf/argmining/KapadnisPPMN21,DBLP:conf/argmining/FriedmanDHAKS21}. However, they are time-consuming, costly and difficult to replicate. Some of the work attempts to use automated evaluation methods such as ROUGE, a metric widely used to evaluate automatically generated summaries \citep{lin2004rouge}. This type of automatic metric compares generated sentences with gold standard ones, but it is difficult to measure their accuracy and effectiveness in terms of capturing semantic similarity. Recent trends consider automated evaluation as different tasks.  \citet{DBLP:conf/iclr/ZhangKWWA20} proposed unsupervised matching metrics, aimed at measuring semantic equivalence by mapping candidates and references to a distributed representation space. \citet{DBLP:conf/acl/SellamDP20} presented a supervised learning evaluation metric that can model human judgments by a novel pre-training  scheme. Their work demonstrates that pre-training a metric on task-specific synthetic data, before fine-tuning it on handpicked human ratings can improve metric robustness. \citet{DBLP:conf/emnlp/ReiSFL20} considered the problem as a ranking task, leveraging breakthroughs in multilingual pre-trained models to generate ratings that resemble human judgments. \citet{DBLP:conf/nips/YuanNL21} instead suggested that evaluating the quality of summaries can be treated as a text generation task. The main idea is that converting a well-performing generated text to/from a reference text would easily achieve higher scores. While these approaches have advanced the field, they all focus on sentence-level evaluation. Our task, however, requires the  evaluation of a set of key points. The reason is that when comparing generated key points to gold-standard annotations at a sentence level, important information could be lost. This can only be retained by considering all sentences at once. 

\section{Methodology}

In this section, we describe our framework in detail. 
As can be seen from Figure \ref{overview}, for each debate topic such as \emph{``Should we abandon the use of school uniforms?''}, we take a corpus of relevant arguments grouped by their stance towards the topic (i.e. ``pro'' or ``con'') as input, as mined from online discussion boards.
As part of KPM, these arguments are clustered using a neural topic modelling approach to group them by their common theme. The clusters are then used as input to the KPG model for summarisation, which is optimised to generate a key point for each argument cluster.
During the training of our model for  KPM, we employ data augmentation. 


\subsection{Key Point Modelling (KPM)}

In previous work, researchers made the simplifying assumption that each argument can be mapped to a single key point \citep{DBLP:conf/argmining/AlshomaryGSHSCP21, DBLP:conf/argmining/KapadnisPPMN21}. As a consequence, finding this mapping was modelled as a classification task. 
In practice, however, a single argument may be related to multiple key points. 
For instance, the argument: ``\emph{School uniforms stifle freedom of expression; they can be costly and make circumstances difficult for those on a budget.}'' expresses the key points ``\emph{School uniform is harming the student's self expression.}'' and ``\emph{School uniforms are expensive.}''.
Inspired by this observation, we approach KPM as \emph{clustering}, by grouping together similar arguments. This naturally allows us to map arguments to multiple key points. Unlike key point matching using a classifier, this step can be performed without any labelled data, since clustering is an unsupervised technique. If training data in the form of argument-key point mappings is available, it is desirable to incorporate this information, as latest work shows that supervision can improve clustering performance \citep{DBLP:conf/ictai/EickZZ04}.
To that end, we use \texttt{BERTopic} as our clustering model \citep{DBLP:journals/corr/abs-2203-05794}, which facilitates the clustering of sentences based on their contextualised embeddings obtained from a pre-trained language model \citep{DBLP:conf/emnlp/ReimersG19}, as well as fine-tuning them further for the clustering task. We convert the key points into numbers as labels for training; arguments that do not match any key points are dropped.



A common challenge of clustering algorithms is the difficulty of clustering data in high-dimensional space. Although several methods to overcome the curse of dimensionality were proposed recently \citep{DBLP:journals/tkdd/PandoveGR18}, the most straightforward way is to reduce the dimensionality of embeddings \citep{DBLP:conf/grapp/MolchanovL18}. We achieve this by applying UMAP on the raw embeddings \citep{DBLP:journals/corr/abs-1802-03426} to reduce their dimension while preserving the local and global structure of embeddings.  
HDBSCAN \citep{DBLP:journals/jossw/McInnesHA17} is then used to cluster the reduced embeddings. 

The output of this step is a set of clusters and the probability distribution of each argument belonging to each cluster. Based on this, we discretise the probability distribution, i.e. represent each argument-cluster pair as a value, which allows us to map arguments to multiple clusters; the formulae and details can be seen in Appendix~B.2. As shown in Figure \ref{overview}, these clustered arguments serve as input for the Key Point Generation model. 

\begin{algorithm}[t!]
\caption{KPM with Iterative Clustering}
\label{alg:algorithm-IC}
\textbf{Input}: Clusters $C$; Unclassified Arguments $Arg$ 
\textbf{Parameter}: Threshold $\lambda$\\
\textbf{Output}: Algorithm Result $IC$

\begin{algorithmic}[1] 
\STATE $IC$ $\leftarrow$ $C$, $\phi$ $\leftarrow$ 0, $l$ $\leftarrow$ len($Arg$), $\omega$ $\leftarrow$ len($C$)
\FOR{$i$ to $l$} 
    \FOR{$J$ to $\omega$} 
        \STATE  $\beta$ $\leftarrow$ compute anchor of $IC$
        \STATE  $\phi$ $\leftarrow$ compute similarity ($a_{i}$,$\beta$) 
        \IF{ $\phi$ $>$ $\lambda$}
            \STATE $IC_{j}$ $\leftarrow$ $IC_{j}$ + $a_{i}$
        \ELSE
            \STATE $IC_{\omega+1}$ $\leftarrow$ $a_{i}$, $C_{\omega+1}$ $\leftarrow$ $a_{i}$
        \ENDIF
        \STATE \textbf{update} $IC$
    \ENDFOR
\ENDFOR

\STATE \textbf{return} $IC$
\end{algorithmic}
\end{algorithm}

\subsection{Iterative Clustering (IC)}
The output of KPM includes a set of arguments that are unmatched, i.e., not assigned to any cluster, represented as a cluster with the label ``-1'', because HDBSCAN  is a soft clustering approach that does not force every single node to join a cluster \citep{DBLP:journals/jossw/McInnesHA17}. 
In order to increase the ``representativeness'' of generated KPs, it is reasonable to maximise the number of arguments in each cluster. To this end, we propose an iterative clustering algorithm (formally described in Algorithm~\ref{alg:algorithm-IC}) to further assign these unmatched arguments according to their semantic similarity to cluster centroids. We compute the semantic similarity between each unclassified argument and the cluster centre, by calculating the vector product of embeddings and the average of clusters.


To tackle the issue of determining the cluster centers, we employ two different techniques: one is by calculating the similarity of the candidates to each sample in the cluster and then taking the average distance, while the other is by taking the centroid of each cluster as the \emph{anchor} \citep{wang2021more}. As a filtering step, each unmatched argument is compared to the anchor. We only assign the argument to the cluster if the similarity is higher than a hyper-parameter $\lambda$; otherwise we create a new cluster. 
Next, the clusters are updated at each iteration until all arguments have been assigned to a cluster. 

\subsection{Key Point Generation (KPG)}

We model KPG as a supervised text generation problem. The input to our model is as follows: \{Stance\} \{Topic\} \{List of Arguments in Cluster\}\footnote{For example: \emph{Positive We should abandon the use of school uniforms. School uniforms are expensive and place an unnecessary burden on the parents of students...}}, where the order of arguments in the list is determined by \texttt{TextRank} \citep{mihalcea2004textrank}. We train the model by minimising the cross-entropy loss between generated and reference key points. The reference key points are drawn from a KPM dataset, together with their matched arguments, which serve as the input to the model.

During inference, we use the list of arguments as provided by KPM as input. The generated KPs are ranked in order of relevance using \texttt{TextRank} \citep{mihalcea2004textrank}. Duplicate KPs with a cosine similarity threshold above $0.95$ are combined 
and the final list of KPs is ranked based on the size of their clusters (for example, the yellow key point with six arguments is ranked higher than the pink key point with four arguments in Figure~\ref{overview}). For combined KPs, we take the sum of the respective cluster sizes.  



\subsection{Data Augmentation (DA)}
Many problems lack annotated data to fully exploit supervised learning approaches.
For example, the popular KPA dataset \textbf{ArgKP-2021} \citep{DBLP:conf/acl/Bar-HaimEFKLS20} features an average 150 arguments per topic, mapped to 5-8 KPs. We rely on data augmentation to obtain more KPM training samples. Specifically, we use DINO \citep{DBLP:conf/emnlp/SchickS21a} as a data augmentation framework, that leverages the generative abilities of pre-trained language models (PLMs) to generate task-specific data by using prompts. 
We customised the prompt for DINO to include task descriptions (i.e., ``\emph{Write two claims that mean the same thing}'') to make the model generate a new paraphrase argument. We then used BERTScore \citep{DBLP:conf/iclr/ZhangKWWA20} and BLEURT \citep{DBLP:conf/acl/SellamDP20} to assess the difference in quality between each generated sample and the corresponding reference, removing 25\% of the lowest scoring generated arguments.

\subsection{Set-level KPG Evaluation}
Other tasks with sets of predictions, such as information retrieval, are evaluated by means of precision and recall, where a set of predictions is compared against a set of references. Since the final output of KPG and the reference KPs are sets, it is desirable to follow a similar evaluation method. However, it is not sufficient to rely on traditional precision and recall, as these are based on direct sentence equivalence comparisons whereby predictions and references might differ in wording although they are semantically similar. Instead, we rely on \emph{semantic similarity measures} that assign continuous similarity scores rather than equivalence comparison to identify the best match between generated and reference KPs---we call these metrics \emph{Soft-Precision} ($sP$) and \emph{Soft-Recall} ($sR$). More specifically, for $sP$, we find the reference KP with the highest similarity score for each generated KP and vice-versa for $sR$. We further define \emph{Soft-F1} ($sF1$) as the harmonic mean between $sP$ and $sR$.


The final $sP$ and $sR$ scores is the average of these best matches. Formally, we compute $sP$ (and $sR$ analogously) as follows:

\begin{equation}
    sP = \frac{1}{n} \times \sum_{ \alpha_i\in\mathcal{A}} \max_{\beta_j\in\mathcal{B}} f(\alpha_{i},  \beta_{j})
\end{equation}
\begin{equation}
    sR = \frac{1}{m} \times \sum_{ \beta_i\in\mathcal{B}} \max_{\alpha_j\in\mathcal{A}} f(\alpha_{i},  \beta_{j})
\end{equation}
where, $f$ computes similarities between two individual key points, $\mathcal A$, $\mathcal{B}$ are the set of candidates and references and $n=|\mathcal{A}|$ and $m=|\mathcal{B}|$, respectively. When $i$ iterates over each candidate, $j$ iterates over each reference and selects the pair with the highest score as the reference for that candidate.

We have chosen state-of-the-art semantic similarity evaluation methods such as BLEURT \citep{DBLP:conf/acl/SellamDP20} and BARTScore \citep{DBLP:conf/nips/YuanNL21} as $f_{max}$. 

\subsection{Implementation Details}

\textbf{KPM with Iterative Clustering}: We first experimented with thresholds at $0.2$ intervals respectively, but the results showed little variation in downstream KPA performance on ROUGE when the threshold was less than $0.6$. Therefore, we compare the influence of  key point quality on ROUGE when the threshold was greater than $0.6$ with 0.1 intervals.
Preliminary experiments showed that cluster sizes vary in length and contain irrelevant or incorrectly assigned arguments. Following the intuition that important sentences should be considered first by the KPG model, we order the input sentences based on their \emph{centrality} in the cluster. Specifically, we use \texttt{TextRank} \citep{mihalcea2004textrank}, such that  sentences receive a higher ranking if they have a high similarity score to all other sentences.

\textbf{Key Point Generation}: 
We choose \texttt{Flan-T5} \citep{DBLP:journals/corr/abs-2210-11416} as our KPG model, which is fine-tuned on more than 1000 different tasks, and it has received a lot of attention as a potential alternative of \texttt{GPT-3} \citep{DBLP:conf/nips/BrownMRSKDNSSAA20}. To maintain comparability to previous work, we only keep $n$ generated key points, where $n$ is the number of key points in the reference. 

\textbf{Data Augmentation}: We employ \texttt{GPT2-XL} \citep{radford2019language} as the data augmentation model with default settings, setting the maximum output length to $40$. Finally, the arguments are matched with the corresponding key points, stance and topics to create a training set of $520$k instances. 
Example templates and the full dataset description can be found in Appendix~A.

\section{Experimental Setup}
Broadly speaking, we aim to investigate the efficacy of our proposed KPM framework as well as the evaluation metrics. Specifically, we ask: 
\emph{(i)} Does the proposed approach improve the performance of the task? \emph{(ii)} Does data augmentation help with the lack of training data? \emph{(iii)} Does the re-clustering of outliers by using IC improve performance on downstream tasks? \emph{(iv)} Does the proposed evaluation framework correlate better with human judgments than raw ROUGE scores? 
To answer question \emph{(i)} we compare the performance of our proposed approach to established previous approaches on the task of KPA. For questions (\emph{ii}) and (\emph{iii}), we perform ablation studies to measure the impact  our using supervised and unsupervised KPM pipelines (\emph{S-KPM} and \emph{US-KPM}) as well as data augmentation (\emph{+DA}) and iterative clustering (\emph{+IC}). For question \emph{(iv)}, we conduct manual evaluation.

\textbf{Baselines}: We compare our approach with previous known and open-source work---Enigma~\citep{DBLP:conf/argmining/KapadnisPPMN21} and Graph-based Summarization (GBS)~\citep{DBLP:conf/argmining/AlshomaryGSHSCP21}\footnote{Note that only key point matching is described in their published paper, but their key point generation code can be found on Github at \url{https://github.com/manavkapadnis/Enigma_ArgMining}}, selecting their best reported results as the baseline. Enigma uses an abstract summarisation approach, employing \texttt{PEGASUS} \citep{DBLP:conf/icml/ZhangZSL20} as the summarisation model, to generate candidate KPs by taking a single argument and its corresponding topic as input. Finally, the top-$n$ highest ROUGE scores with reference KPs were selected as the final result. Similar to the work of \citet{DBLP:conf/sigir/AlshomaryDW20}, GBS constructs an undirected graph with arguments as nodes. Nodes with sufficiently high argument quality scores \citep{DBLP:conf/emnlp/ToledoGCFVLJAS19}, and node matching scores \citep{DBLP:conf/argmining/AlshomaryGSHSCP21} are connected. The importance score of each argument is then calculated based on \texttt{PageRank} \citep{page1999pagerank} and ranked in descending order. Finally,  only those arguments where matching scores are below the threshold of the already selected candidates are added to the final set of key points.

\textbf{Evaluation metrics}: We calculate ROUGE~\citep{lin2004rouge} scores on the test set, by comparing the concatenation of all generated key points to the concatenation of the reference, averaging for all topic and stance combinations. Furthermore, in order to evaluate the quality of the generated key points invariant to the order of sentences, we also compare the performance based on the proposed set-level evaluation approach. Similar to our idea, the earth mover's distance (EMD)~\citep{rubner2000earth} is a measure of the similarity between two data distributions. By combining Word Mover's Distance (WMS) \citep{DBLP:conf/icml/KusnerSKW15} and Sentence Mover's Similarity (SMS) \citep{DBLP:conf/acl/ClarkCS19}, Sentence + Word Mover's Similarity (S+WMS) measures both the word distribution of a single sentence and similarity at the set level. However, an observable shortcoming is that they consider a set of sentences as a single paragraph, without splitting and using GloVe embeddings \citep{DBLP:conf/emnlp/PenningtonSM14} instead of fine-tuning on sentence-level similarity.

\begin{table*}[t]
    \centering
    \scalebox{1}{
    \begin{tabular}{lcccccccccc}
    \toprule
    {} & \multicolumn{3}{c}{ROUGE} & \multicolumn{3}{c}{BLEURT} & \multicolumn{3}{c}{BARTScore}\\
    \cmidrule(r){2-4} \cmidrule(r){5-7} \cmidrule(r){8-10} 
    ${\rm Approach_{Size}}(Setting)$ & R-1 & R-2 & R-L & sP & sR & sF1 & sP & sR & sF1 & S+WMS\\
    \hline

    \emph{${\rm SKPM_{11B}}(DA+IC)$} & \textbf{32.8} & \textbf{9.7} & \textbf{29.9} & \textbf{0.70} & \textbf{0.71} & \textbf{0.71}  & \textbf{0.73} & \textbf{0.79} & \textbf{0.76} &  \textbf{0.0416}\\
    \emph{${\rm SKPM_{3B}}(DA+IC)$} & 32.2 & 9.0 & 27.9 & 0.68 & 0.67 & 0.67 & 0.58 & 0.71 & 0.64 & 0.0382\\
    \emph{${\rm SKPM_{Large}}(DA+IC)$} & 31.4 & 9.1 & 28.1 & 0.57 & 0.62 & 0.60 & 0.54 & 0.75 & 0.63 & 0.0276 \\
    \emph{${\rm SKPM_{Base}}(DA+IC)$} & 30.3 & 8.9 & 28.1 & 0.59 & 0.58 & 0.59 & 0.57 & 0.63 & 0.60 & 0.0320\\
    \emph{${\rm SKPM_{Base}}(DA)$} & 30.7 & 9.1 & 27.6 & 0.58 & 0.58 & 0.58 & 0.53 & 0.66 & 0.59 & 0.0304 \\
    \emph{${\rm SKPM_{Base}}(IC)$} & 28.9 & 9.2 & 28.3 & 0.62 & 0.57 & 0.59 & 0.53 & 0.60 & 0.57 & 0.0332\\
    \emph{${\rm SKPM_{Base}}$} & 24.9 & 6.1 &24.0 & 0.55 & 0.55 & 0.55 & 0.53 & 0.67 & 0.59 & 0.0279\\
    \emph{${\rm USKPM_{Base}}(IC)$} & 29.5 & 7.8 & 28.1 & 0.61 & 0.57 & 0.59 & 0.54 & 0.66 & 0.60 & 0.0318\\
    \emph{${\rm KMeans_{Base}}$} & 26.5 &7.3 &25.5 & 0.59 & 0.56 & 0.57 & 0.53 & 0.69 & 0.60 & 0.0264 \\
    \emph{${\rm USKPM_{Base}}$} & 25.2 & 5.7 & 23.2 & 0.59 & 0.53 & 0.56 & 0.52 & 0.63 & 0.57 & 0.0306\\
    \hdashline
    \emph{Enigma} & 20.0 & 4.8 & 18.0 & 0.58 & 0.57 & 0.57 & 0.54 & 0.69 & 0.61 & 0.0368\\
    
    \emph{GBS} (Baseline) & 19.8 & 3.5 & 18.0 & 0.51 & 0.54 & 0.53 & 0.53 & 0.66 & 0.59 & 0.0258\\
    \emph{GBS} (Ours) & 19.6 & 3.4 & 17.7 & 0.53 & 0.52 & 0.52 & 0.53 & 0.71 & 0.61 & 0.0250\\
    \emph{Aspect Clustering} & 18.9 & 4.7 & 17.1 & - & - & - & - & - & - & - \\
    \bottomrule
    \end{tabular}
    }
    \caption{\label{table:corpus-level evaluation}
    Test set ROUGE scores and the proposed set-based evaluation metrics. Soft-Precision, Soft-Recall, and Soft-F1 are reported using BLEURT and BARTScore as $f_{max}$. GBS is a graph-based summarisation method and GBS (Ours) is the result when the number of references is the same as that generated. The results of Aspect Clustering are reported directly from the paper \citep{DBLP:conf/argmining/AlshomaryGSHSCP21}, as their code is not open source. $11B$, $3B$, $Large$ and $Base$ refer to \texttt{Flan-T5-xxl}, \texttt{Flan-T5-xl}, \texttt{Flan-T5-Large} and \texttt{T5-effictive-base}, respectively. S+WMS stands for Sentence + Word Mover's Similarity \citep{DBLP:conf/acl/ClarkCS19}.
    }
\end{table*}

\textbf{Human Evaluation}: Taking into account the wealth of problems arising from automatically evaluating generated texts, we further verify the reliability of our obtained results,by means of human evaluation. Seven annotators were selected, all of whom are graduate students with a diploma from a University in the UK. Before starting, all annotators received task-oriented training, the specific instructions can be found in Appendix~C.1. After an introduction, they had to answer a questionnaire containing $66$ questions for all topics and stances in the test set. The annotators were asked to answer on a Likert scale ranging from ``very good'' ($5$) to ``not good at all'' ($1$).

The first evaluation task (HT1) investigates how well the generated summaries of clusters serve as KPs. Following \citet{DBLP:conf/acl/Bar-HaimEKFS20}, we assessed the quality of the key points in four requirements: \textsc{Validity}, \textsc{Sentiment}, \textsc{Informativeness} and \textsc{SingleAspect}. Annotators are asked to read three sets of KPs separately (reference, our best approach, previous work), assigning each of the four dimensions above a single score, and then ranking each of the three outputs the outputs from best to worst.

The second task (HT2) evaluates how well the generated set of key points summarises the corpus of arguments. In previous work crowdworkers evaluated how well generated key points represent a given corpus of arguments \citep{DBLP:conf/argmining/FriedmanDHAKS21}. However, they only considered \textsc{Redundancy} and \textsc{Coverage}, as the outputs key points were extracted from a corpus, rather than generated. To adapt their experiment to the generative setting, We additionally define \textsc{Significance} (i.e. how well a KP uniquely captures a theme) and \textsc{Faithfulness} (i.e. no unfounded claims are conjectured). 
We refer the reader to Appendix~C.2 for the full definition of all quality dimensions.

Finally, in the third evaluation task (HT3), we investigate how well automated evaluation metrics correlate with human judgement. Here, the annotators were asked to perform pair-wise comparison between two sets of generated KPs for which the difference between ROUGE scores and the soft-F1 metric was the highest.

\section{Results and Analysis}

\begin{table}
\centering
\begin{tabular}{ccccccc}
   \toprule
   Threshold & 0.6 & 0.7 & 0.8 & 0.9 \\
   \midrule
   R-1 & 25.5 & 27.7 & 28.9 & 29.1 \\
   R-2 & 6.0 & 5.9 & 6.4 & 7.5 \\
   R-L & 24.3 & 25.9 & 27.0 & 27.2 \\
   \bottomrule
\end{tabular}
\caption{ROUGE for different threshold values on IC}
\label{ROUGE_threshold}
\end{table}

\begin{table*}[t]
    \centering
    \scalebox{0.89}{
    \begin{tabular}{p{2.7cm}p{6.6cm}p{7.4cm}}
    \toprule
    (Stance) Topic  & \emph{Sup-KPM+DA}  & \emph{Unsup-KPM+IC}\\
    \midrule
    (Con) Routine child vaccinations should be mandatory & (1) The Routine child vaccinations should not be mandatory. (2) The parents should decide for their child. (3) The vaccine can cause harm to the child. & (1) Child vaccinations should not be mandatory because many times children cannot catch the virus. (2) Parents should have the freedom to choose what is best for their child. (3) Child vaccinations can lead to harmful side effects.\\
    \hdashline
    (Pro) Social media platforms should be regulated by the government & (1) Social media platforms can be regulated to prevent terrorism. (2) Social media platforms should be regulated to prevent hate crimes. (3) Social media platforms can be regulated to prevent spreading of false news. & (1) Social media platforms should be regulated to prevent rumors/harming the economy. (2) Social media platforms should be regulated to prevent hate crimes. (3) Social media platforms should be regulated to prevent inappropriate content.\\
    \bottomrule
    \end{tabular}
    }
    \caption{Examples of generated KPs from proposed approach. For the sake of brevity, only the top three key points are shown.}
    \label{tab:example_kpg}
\end{table*}

\begin{table}[b]
\centering
\begin{tabular}{ccc}
   \toprule
   Methods	& R-value & P-value \\
   \midrule
   Rouge & 0.61 & 0.03 \\
   Soft-F1 & 0.72 & 0.01 \\
   S+WMS & 0.60 & 0.04 \\
   \bottomrule
\end{tabular}
\caption{Spearman’s correlation between human-assigned scores and the metrics ROUGE, soft-F1 and EMD. The inputs used in the calculations are only those systems included in the human evaluation.}
\label{Spearman}
\end{table}

\textbf{Proposed approach improves performance on KPA task}: Our proposed two-step method outperforms the reference implementations on the full KPA task, with improvements of up to $12\%$ and $14\%$ in ROUGE and Soft-F1, respectively, as shown in Table \ref{table:corpus-level evaluation}.

Overall, each proposed improvement (\emph{+DA} and \emph{+IC}) contributes to achieve better scores. A robustness experiment was then performed on the best-performing approach, with 10 runs, showing that the overall performance is still up to 11\% superior compared to the baseline according to ROUGE, and up to 3\% superior based on the proposed evaluation approach.

It is worth noting that unsupervised KPM with IC (\emph{US-KPM+IC}) yields increases of more then ten points in ROUGE-L and two soft-F1 (BLEURT) percent points compared to the best performing baseline, demonstrating that the proposed method outperforms previous state-of-the-art approaches even without training the clustering model and relying on data augmentation. Our human evaluation further supports these findings: in the ranking task T1, our method was ranked higher than the baselines, slightly behind human-written reference KPs. As can be seen from Table~\ref{human_result}, the annotators consider our work to be slightly worse ($4.5$) than the gold standard in terms of \textsc{Sentiment}, but comparable in performance on the other dimensions (between $4.5$ and $4.7$). In comparison to other work, our approach outperforms the baseline in all dimensions. This is especially significant for \textsc{Coverage} ($4.6$ vs $4.0$) and \textsc{Redundancy} ($4.5$ vs $3.2$), as it suggests that our approach to KPA better captures unique themes across the corpus of arguments and effectively reduces redundancy in the KPs. It is worth noting that annotators generally preferred it when the output consisted of a few general KPs (\emph{Ref}, \emph{S-KPM+IC+DA}) rather than a higher number of specific ones (\emph{GBS}). This contradicts the conclusion made by \citet{DBLP:conf/acl/SyedKAWP21}. However, they suggested summarising long texts into a single conclusion, whereas we focused on summarising a body of short texts (i.e. arguments) in terms of multiple key points.

\textbf{Data augmentation helps}: In the ablation experiments, data augmentation in the supervised scenario shows a significant improvement (\emph{S-KPM-DA} vs. \emph{S-KPM}), by around $4$ points on ROUGE-L and up to $3$ points on proposed evaluation metrics. A possible reason for this improvement is likely because the original dataset is too small for supervised models to learn task-specific representations. Employing prompt-based data augmentation leverages the pre-training of language models, by aligning the down-stream task (i.e. generating similar data) to the pre-traing task~\citep{DBLP:journals/corr/abs-2107-13586}. As a consequence, the proposed data augmentation method can generate training data of sufficient quality to improve downstream KPM performance, even after training the DA model with only a limited amount of annotated data.

\textbf{IC improves the clustering performance}: 
For unsupervised KPM,  iterative clustering (\emph{US-KPM+IC}), performs significantly better than the method with no such additional processing step  (\emph{US-KPM}), showing an increase of $5$ points in terms of ROUGE-L. The gap closes for supervised models (\emph{S-KPM}), presumably due to the fact that after supervision, the KPM model produces less outliers to be further assigned with IC. 
Furthermore, Table~\ref{ROUGE_threshold} demonstrates the relationship between the threshold and the performance of the model. There is a strong positive correlation---increasing the threshold results in higher ROUGE scores (Spearman's $r=0.94, p=2.5e^{-9}$). We further implemented an ablation experiment to compare the performance of K-Means and HDBSCAN in order to investigate the research question of whether the IC step may be unnecessary if a different clustering method was applied to the reduced embeddings. The results show that K-Means performs better than Unsup-KPM ($ROUGE=25.5, sF1=0.57$ vs. $ROUGE=23.2, sF1=0.56$) but worse than Unsup-KPM+IC ($ROUGE=28.1, sF1=0.59$). This supports our hypothesis that the arguments labelled as ``-1'' are meaningful. K-Means assigns them to an existing cluster which is better than discarding them completely (KPM without IC), while IC is more accurate in finding (potentially new) clusters for them. It also demonstrates that the proposed iterative distance is useful.

\textbf{The proposed evaluation framework better reflects human judgement}: We note several important differences between our proposed metrics and ROUGE-based evaluation. For instance, \emph{S-KPM+DA} has higher ROUGE scores than \emph{Unsup-KPM+IC}, while \emph{Unsup-KPM+IC} performs worse than \emph{S-KPM+DA} according to both Soft-F1 and human evaluation. One possible explanation is that ROUGE focusses on the overlap of $n$-grams rather than on semantic similarity, resulting in the fact that summaries that repeat words appearing in the reference, but with a lower semantic similarity overall, may receive higher scores. Table~\ref{tab:example_kpg} exemplifies this assumption, as KPs generated by \emph{S-KPM+DA} are less informative and more concise than those generated by \emph{US-KPM+IC}. When directly comparing two sets of KPs produced by \emph{Sup-KPM+DA} and \emph{Unsup-KPM+IC} (HT3), $80\%$ of the annotators indicated that as a whole, the \emph{US-KPM+IC} outperforms \emph{S-KPM+DA}. The remaining $20\%$ consdered both to be of equal quality. In addition, we conducted supplementary experiments to investigate the difference with existing methods. Similar to our set-based method, \citet{DBLP:conf/acl/ClarkCS19} evaluated texts in a continuous space using word and sentence embeddings (S+WMS). As shown in Table \ref{table:corpus-level evaluation}, the proposed methods are higher than the baseline by 5 points, emphasising the superiority of our approach. To further substantiate the claim that our proposed metrics better correlate with human judgement than the prevalent methodology based on ROUGE and S+WMS, we investigate Spearman's \citep{akoglu2018user} correlation between human-assigned scores (averaged for all dimensions) and the metrics ROUGE ($r_{ROUGE}$), soft-F1 ($r_{sF1}$) and S+WMS ($r_{S+WMS}$), for all evaluated models and test set topics. Table \ref{Spearman} demonstrates our finding that Soft-F1 is indeed a more truthful reflection of human judgment than ROUGE ($r_{sF1}=0.72, p=0.01$ vs. $r_{ROUGE}=0.61, p= 0.03$) and S+WMS ($r_{sF1}=0.72, p=0.01$ vs. $r_{S+WMS}=0.60, p= 0.04$). 


\begin{table}
    \centering
    \scalebox{0.7}{
    \begin{tabular}{lcccc:cccc}
         \toprule
         Approach & VL & SN & IN & SA & SG & CV & FF & RD \\
         \midrule
         \emph{Reference} & 5.0 & 4.9 & 4.9 & 4.9 & 4.6 & 4.9 & 4.8 & 4.9 \\
         \emph{S-KPM+DA+IC} & 4.7 & 4.5 & 4.6 & 4.7 & 4.2 & 4.6 & 4.6 & 4.5 \\
         \emph{S-KPM+DA} & 4.8 & 4.4 & 3.4 & 3.0 & 3.2 & 4.4 & 3.4 & 2.7\\
         \emph{US-KPM+IC} & 4.9 & 4.9 & 4.5 & 4.3 & 4.1 & 4.6 & 4.5 & 4.0\\
         \hdashline
         \emph{Enigma} & 4.6 & 4.2 & 3.0 & 2.5 & 2.7 & 4.0 & 3.0 & 2.2 \\
         \emph{GBS} & 4.7 & 4.3 & 4.7 & 3.5 & 4.0 & 3.9 & 3.7 & 3.2 \\
         \bottomrule
    \end{tabular}
    }
    \caption{Performance of different approaches on each dimension in human evaluation. Each score is averaged over seven annotators on the dimension (HT1 and HT2 are on the left and right of the vertical broken line, respectively). Reported are, from left to right, \textsc{Validity}, \textsc{Sentiment}, \textsc{Informativeness}, \textsc{SingleAspect}, \textsc{Significance}, \textsc{Coverage}, \textsc{Faithfulness} and \textsc{Redundancy}}
    \label{human_result}
\end{table}

\textbf{Human evaluation is reliable}: We measured Krippendorff's $\alpha$ \citep{hayes2007answering} to investigate inter-annotator agreement, reporting an average of $0.61$ across all test set topics and quality dimensions, implying that the results are moderately reliable.
The human evaluation is more reliable for \textsc{Sentiment}, \textsc{SingleAspect} and \textsc{Redundancy} with $\alpha$ of $0.69$, $0.69$ and $0.74$, respectively. One possible explanation is that these dimensions are dichotomous,  and thus are more likely for annotators to produce definite results---for example \textsc{Sentiment} measures whether KPs have a clear stance towards the topic, while \textsc{Redundancy} essentially asks whether KPs are duplicated. Conversely, reliability scores are lower for \textsc{Significance} and \textsc{Faithfulness} ($\alpha= 0.53$ for both), likely because these dimensions are susceptible to annotator bias and rely on their knowledge. For example, \textsc{Faithfulness} measures how well the KPs reflect arguments in the corpus. This requires annotators to have a good understanding of the debate topic 
which might be difficult to achieve in practice. 
Evaluation scores and agreements for all dimensions and test set topics are in Appendix~C.3.

\section{Conclusion}

This paper contributes to the development of key point analysis. Firstly, we proposed a two-step abstractive summarisation framework. Compared with previous work, our approach achieves performance on par with a human without additional training samples. Secondly, we developed a new evaluation toolkit, whose effectiveness was demonstrated with human annotators, presenting a more realistic view of the generated KPs' quality than traditional automatic evaluation  metrics. In future work, we will address the issue that KPs with few matching arguments are difficult to cluster, by using contrastive learning \citep{DBLP:conf/naacl/ZhangNWLZMNAX21} to facilitate better intra-cluster and inter-cluster distances.

\section*{Limitations}

Recruiting human subjects for annotation limits the reproducibility of human evaluation. In addition, we have only tested the performance of the proposed framework on the fixed dataset, ArgKP-2021, that we described above, and not on a wider range of data. This is because ArgKP-2021 was the only dataset available for use in this task. Finally, we did not filter the arguments in the original corpus, with the result that potentially offensive arguments may come into the framework as input and generate key points which some readers might find offensive. It is worth noting, however, that the identification of offensive language is not the aim of this work.

\section*{Ethics Statement}

For the present work, we used an existing anonymised dataset without any data protection issues. In addition, all annotators were systematically trained and explicitly informed that their work would be used in the study before human evaluation. The annotators' work was only taken into account if they clearly understood the task and consented to how their work will be used. In addition, we do not collect their names or personal information, only their ratings. Therefore, institutional ethical approval was not required.

\section*{Acknowledgements}
We thank the anonymous reviewers from the ARR December 2022 cycle for their valuable feedback. We would also like to acknowledge the use of the Computational Shared Facility at The University of Manchester. This work was partially funded
by the European Union’s Horizon 2020 research and innovation action programme, via the AI4Media Open Call \#1 issued and executed under the AI4Media project (Grant Agreement no. 951911).
\bibliography{anthology,acl2023}
\bibliographystyle{acl_natbib}

\newpage
\appendix

\section{Data Augmentation}
\label{sec:appendix}

\subsection{Data Description}

\begin{table}[htbp]
\centering
\scalebox{0.85}{
\begin{tabular}{cccc}
   \hline
   Data Set & Arg & Single Arg-KP & Multiple Arg-KP \\
   \hline
   Train(24) & 5583 & 3778 & 238(2) \\
   Dev(4)  & 932 & 604 & 67(0) \\
   Test(3) & 723  & 454 & 46(6) \\
   \hline
\end{tabular}
}
\caption{\label{Dataset} Data Set Statistics.}
\end{table}

In this work, we use the dataset \textbf{ArgKP-2021},  which contains arguments obtained by crowd-sourcing on 31 topics and key points written by experts \cite{DBLP:conf/argmining/FriedmanDHAKS21}. 27k samples are present in the form of $\langle$~argument,~key~point,~label~$\rangle$ triples, and are grouped by positive or negative stance. Labels are crowd-sourced judments of whether a post is an argument, and which arguments are represented by which key points. Table \ref{Dataset} shows that 5\% of the arguments are matched with multiple key points and 27\% of the arguments do not match any of the key points. The dataset was divided at the topic level, with the training, validation and test subsets corresponding to 24, 4 and 3 topics respectively (where the topics across the subsets do not overlap with each other).As mentioned earlier, only 0.001\% of the arguments (2 out of 238 in the training set, 6 out of 46 in the test set and none in the validation set) matched more than three key points



\subsection{Example of template}

\begin{figure}[htbp]
    \centering
    \includegraphics[width=\columnwidth]{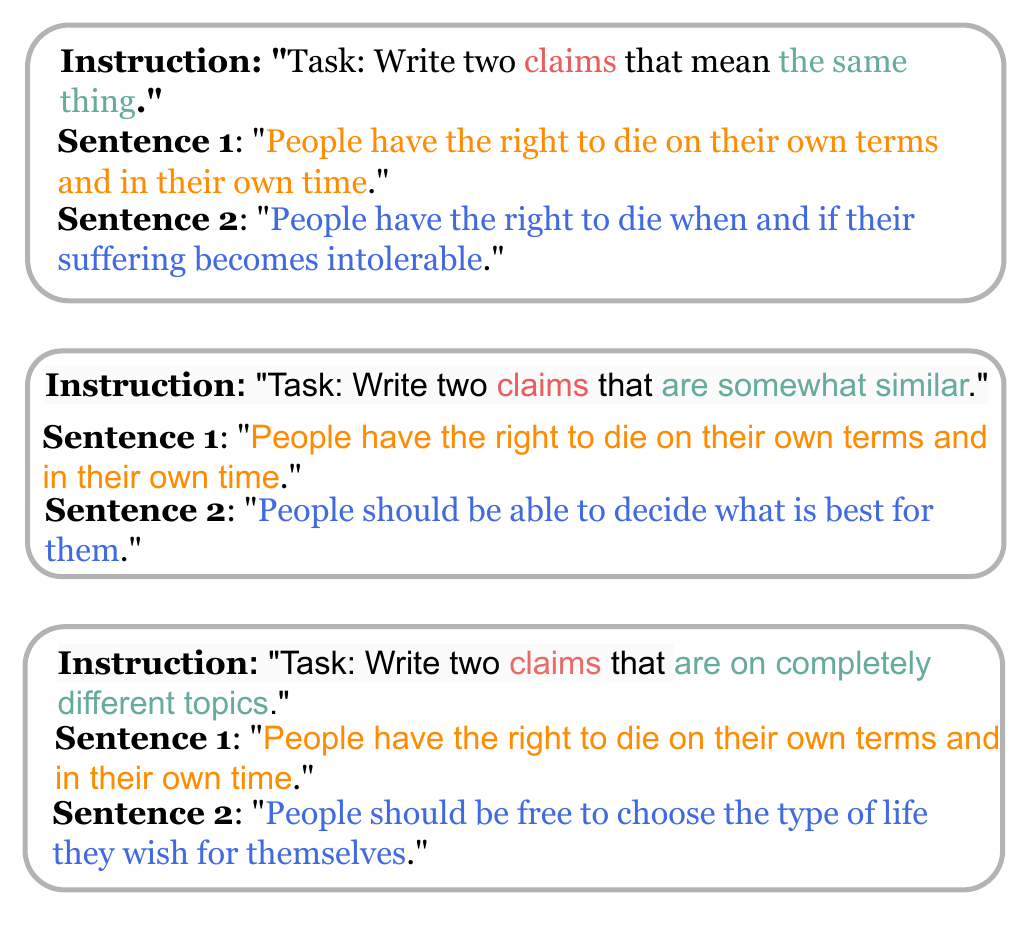}
    \caption{Continuation text generated by prompted learning data augmented methods with three different template descriptions. We chose to give input sentence 1 and generate only sentence 2, which helps to generate sentence similarity datasets.}
    \label{data_aug}
\end{figure}

\subsection{Result of data distribution of the data augmentation dataset}

\begin{figure}[ht]
    \centering
    \includegraphics[width=\columnwidth]{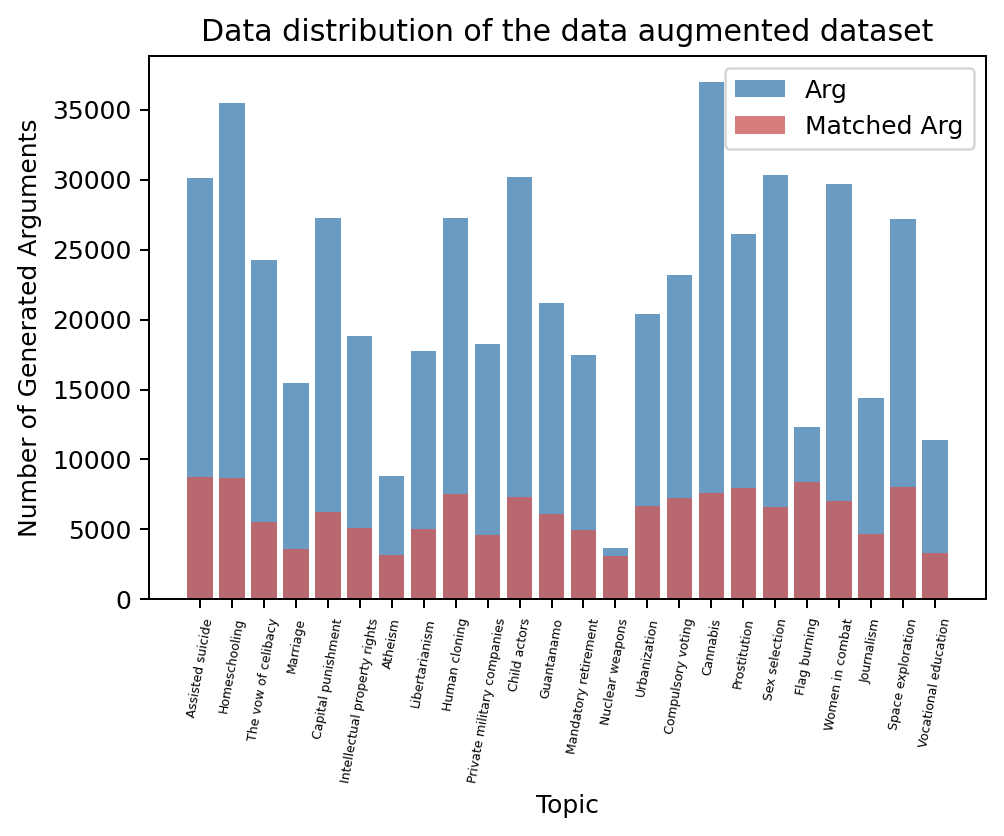}
    \caption{Data distribution of the data augmented dataset}
    \label{data_dis}
\end{figure}

Figure \ref{data_dis} illustrates the data distribution of the final augmented dataset, with each topic containing an average of 20,000 arguments and 7,500 arguments matched to key points.

\section{More details of the methodology}

\subsection{Parameters for DA}

We set DINO's num entries per input and label to 50 which generates 50 data for each label (0, 0.5, 1) of each input example, top p to 0.9, top k to 5 and other parameters follow the default. The DINO \citep{DBLP:conf/emnlp/SchickS21a} is trained on a single NVIDIA Tesla 32G V100 GPU, with each run taking up to twelve hours. 

\subsection{Filtering mechanism for KPM}

By thresholding the unclassified arguments, we take into account the second highest probability. Formally, this procedure is  described as follows: 


\begin{equation}
    \gamma = \frac{\sum_{i=1}^n P_{second-max}(Arg_i)}{n}
\end{equation}
where $\gamma$ is the value of the threshold, $Arg_i \in Input Text$ is an independent argument, $i$ iterates over the second highest probability of each argument, and $n$ is the number of arguments per stance per topic. 
We average the sum of the second highest probabilities as the threshold for selecting the arguments since only 0.001\% of the arguments matched more than two key points, and the third highest probability was more different from the top two (Data distribution details can be seen in Appendix~A.1).

\begin{table}[t]
    \centering
    \scalebox{0.5}{
    \begin{tabular}{lcccccc}
    \toprule
    {} & \multicolumn{3}{c}{ROUGE} & \multicolumn{3}{c}{BLEURT}\\
    \cmidrule(r){2-4} \cmidrule(r){5-7}
    Approach & R-1 & R-2 & R-L & sP & sR & sF1 \\
    \cmidrule(r){1-1} \cmidrule(r){2-4} \cmidrule(r){5-7}

    \emph{Experiment 1} & 30.7 & 9.1 & 28.3 & 0.61 & 0.59 & 0.60 \\
    \emph{Experiment 2} & 31.4 & 9.3 & 29.0 & 0.62  & 0.59 & 0.61 \\
    \emph{Experiment 3} & 29.7 & 9.8 & 27.9 & 0.57 & 0.55 &  0.56\\
    \emph{Experiment 4} & 30.2 & 9.5 & 28.1 & 0.60 & 0.57 & 0.58 \\
    \emph{Experiment 5} & 31.1 & 8.7 & 28.9 & 0.61 & 0.59 &  0.60\\
    \emph{Experiment 6} & 27.8 & 7.0 & 26.4 & 0.55 & 0.56 &  0.56\\
    \emph{Experiment 7} & 31.1 & 8.7 & 29.0 & 0.61 & 0.58 & 0.60 \\
    \emph{Experiment 8} & 30.1 & 9.5 & 28.2 & 0.60 & 0.56 &  0.58\\
    \emph{Experiment 9} & 30.3 & 8.9 & 28.1 & 0.59 & 0.58 & 0.59 \\
    \emph{Experiment 10} & 31.7 & 8.5 & 29.6 & 0.62 & 0.62 & 0.62 \\
    \hdashline
    \emph{Average} & 29.8$\pm$2 & 8.4$\pm$1.4 & 28.0$\pm$1.6 & 0.59$\pm$0.03 & 0.58$\pm$0.04  & 0.59$\pm$0.03 \\
    \bottomrule
    \end{tabular}
    }
    \caption{\label{table:10 times evaluation}
    10 times running result of our best approaches. The experiments were performed on \texttt{T5-effective-base}.
    }
\end{table}

\subsection{Experimental parameters for KPG}
We train the model for a total of $15$ epochs on two NVIDIA Tesla A100 80GB GPUs with and batch size of $16$, limiting input length to $512$.

\subsection{Second set-based automatic evaluation Design}

Due to their outstanding multiple task-based formulation and ability to utilize the entirety of the pre-trained model's parameters, we propose two different lines to use flexibly in different evaluation scenarios. Specifically, the first consideration is that the number of generated key points is likely to be different from the number of reference key points,  presented as in evaluating them from different directions, which are already explained in the main page.

In addition, we propose an evaluation idea specifically for the scenario where the number of generated key points is the same as the number of reference key points. For n generated and reference key points find n pairs of (generated, reference) with maximum score, such that:
\begin{itemize}
    \item Each generated and reference key point appears in some pair
    \item Each generated and reference key point appears only once
\end{itemize}

\subsection{Result of different methods}

\begin{table*}
    \centering
    \scalebox{0.89}{
    \begin{tabular}{p{3cm}p{1cm}p{5.9cm}p{5.9cm}}
    \toprule
    Topic  & Stance & Threshold 0.6  & Threshold 0.9\\
    \midrule
    The USA is a good country to live in & Pro & (1) United States is the best country to live. (2) The United States has a lot of diversity. (3) USA is the American dream. & (1) United States offers many opportunities. (2) The USA has a good standard of living. (3) The USA offers opportunities for everyone to achieve the American dream.\\
    {}&{}&{}&{}\\
    Social media platforms should be regulated by the government & Con & (1) Social media platforms cannot be regulated by the government. (2) Social media platforms are important to freedom of expression. (3) Private companies should not be regulated. & (1) The social media platforms should not be regulated because they are private companies. (2) Social media platforms should be regulated to prevent crimes. (3) Social media platforms should not be regulated because it would be ineffective.\\
    \bottomrule
    \end{tabular}
    }
    \caption{Examples of key points generated from our proposed approach. For the sake of brevity, only the top three key points are shown.}
    \label{tab:example_kpg_threshold}
\end{table*}
Table \ref{tab:example_kpg_threshold} shows the example generated KPs based on different threshold. Table \ref{table:corpus-level evaluation} demonstrates the different work in sP,sR and sF1 based on BARTScore. Table \ref{table:10 times evaluation} shows the overall performance of \emph{S-KPM+IC+DA} after 10 times running.

\section{Human Evaluation}

\subsection{Tutorial for human evaluation}
The main aim of this evaluation is to assess the quality of the argument summaries automatically generated by the language model. Unlike summaries of articles, this task is presented by a highly condensed set of sentences as a summary. Each of them is known as a key point. Following is an example:

\textbf{Topic}: We should abandon the use of school uniform

\textbf{Stance}: Con

\textbf{Original text}:

1. School uniform keeps everyone looking the same and prevents bullying.

2. Having a school uniform can reduce bullying as students who have no style or cannot afford the latest trends do not stand out.

3. School uniforms can prevent bullying due to economic background and appearance.

\textbf{Key point}: School uniform reduces bullying.

\textbf{Task description}

There are three tasks involved in this evaluation.
The first task concerns how well the summary itself serves as a key point.
The second task aims to determine which of the two sets of generated key points is more consistent with the way humans produce summaries.
The third task evaluates how well the generated set of key points summarises the corpus of arguments.

\subsection{Dimensions of human evaluation}
Annotators were asked to evaluate the gold annotated key points as ground truth, followed by an evaluation of the best performing set of generated key points. Before starting, they were given task-oriented training that explained in detail the definition of arguments, key points and topics. The following are the dimensions involved in the evaluation task.
\begin{itemize}
    \item VALIDITY: The key point should be an understandable, well-written sentence.
    \item SENTIMENT: It should have a clear stance towards the debate topic (either positive or negative).
    \item INFORMATIVENESS: It should discuss some aspect of the debate topic and be general enough. Any key point that is too specific or only expresses sentiment cannot be considered a good candidate.
    \item SINGLEASPECT: It should not involve multiple aspects.
\end{itemize}

\begin{itemize}
    \item SIGNIFICANT: Each key point should stand out and capture a main point.
    \item COVERAGE: A set of KPs should cover the most of semantic information in a given corpus.
    \item FAITHFULNESS: KPs should actually express the meaning in the corpus. No conjecture or unfounded claims arise.
    \item REDUNDANT: Each KP expresses a distinct aspect. In other words, there should be no overlap between the key points.
\end{itemize}

\subsection{Results of human evaluation}

The following table shows the consistency between the human annotators on a different topics.
\begin{table}[htbp]
    \centering
    \scalebox{0.7}{
    \begin{tabular}{lcccc:cccc}
         Topic & VL & SN & IN & SA & SG & CV & FF & RD \\
         \midrule
         \emph{Routine-Con} & 0.46 & 0.56 & 0.49 & 0.84 & 0.42 & 0.75 & 0.49 & 0.79\\
         \emph{Routine-Pro} & 0.62 & 0.62 & 0.64 & 0.54 & 0.33 & 0.62 & 0.48 & 0.68\\
         \emph{Media-Con} & 0.45 & 0.84 & 0.64 & 0.58 & 0.40 & 0.67 & 0.43 & 0.54\\
         \emph{Media-Pro} & 0.29 & 0.63 & 0.46 & 0.52 & 0.50 & 0.54 & 0.35 & 0.73\\
         \emph{USA-Con} & 0.32 & 0.66 & 0.76 & 0.78 & 0.74 & 0.46 & 0.72 & 0.80 \\
         \emph{USA-Pro} & 0.25 & 0.82 & 0.77 & 0.70 & 0.80 & 0.60 & 0.72 & 0.85\\
         \hdashline
         \emph{Average} & 0.40 & 0.69 & 0.63 & 0.69 & 0.53 & 0.61 & 0.53 & 0.74 \\
         \bottomrule
    \end{tabular}
    }
    \caption{Result of Krippendorff's Alpha on each dimension. Each score is the average score of seven annotators on the dimension (HT1 left and HT2 right). Reported are, from left to right, \textsc{Validity}, \textsc{Sentiment}, \textsc{Informativeness}, \textsc{SingleAspect}, \textsc{Significance}, \textsc{Coverage}, \textsc{Faithfulness} and \textsc{Redundancy}}
    \label{human_result_kp}
\end{table}

\end{document}